\documentclass[conference]{IEEEtran}
\IEEEoverridecommandlockouts
\usepackage{cite}
\usepackage{amsmath,amssymb,amsfonts}
\usepackage{algorithmic}
\usepackage{algorithm}
\usepackage{graphicx}
\usepackage{textcomp}
\usepackage{xcolor, soul}
\usepackage{multirow}
\usepackage{mdframed}
\usepackage{amsthm}
\usepackage{framed}
\usepackage{mdframed}
\newtheorem*{remark}{Remark}
\usepackage{subcaption}
\usepackage{booktabs}
\usepackage{xfrac}
\usepackage{hyperref}
\usepackage{efbox}
\usepackage[sort&compress,numbers,square,comma]{natbib}
\def\BibTeX{{\rm B\kern-.05em{\sc i\kern-.025em b}\kern-.08em
    T\kern-.1667em\lower.7ex\hbox{E}\kern-.125emX}}

\newcommand{\comment}[1]{}

\definecolor{mohamed_colour}{RGB}{255, 204, 205}

\comment{
\linespread{0.99}

\addtolength{\textfloatsep}{-15pt}
\addtolength{\dbltextfloatsep}{-10pt}

\addtolength{\floatsep}{-10pt}
\addtolength{\dblfloatsep}{-5pt}
\addtolength{\abovecaptionskip }{-4pt}

}

\begin{document}

\title{Mapping Deep Learning Workloads Onto Heterogeneous Hardware Platforms\\
}

\title{Mapping Deep Learning Workloads Onto Heterogeneous Hardware Platforms\\
}

\title{Mapping Deep Neural Network Inference Onto Heterogeneous Hardware Platforms\\
}

\title{Mapping Deep Neural Networks Onto Heterogeneous Hardware Platforms\\
\vspace{-1cm}
}

\title{Mapping Neural Network Inference Onto Heterogeneous Hardware Platforms\\
\vspace{-1cm}
}

\title{MILP-SPLIT: Mapping Neural Networks Inference onto Heterogeneous Hardware Platforms\\
\vspace{-1cm}
}

\title{Mapping Neural Networks Inference onto Heterogeneous Hardware Platforms\\
\vspace{-1cm}
}

\title{DiviML: Divide and Conquer for Heterogeneous Mapping of Neural Networks with Lower-Bound Estimation\\
\vspace{-1cm}
}

\title{DiviML: Divide and Conquer for Mapping Neural Networks on Heterogeneous Platforms\\
\vspace{-1cm}
}

\title{DiviML: A Module-based Heuristic for Mapping Neural Networks onto Heterogeneous Platforms\\
\thanks{Thanks to TATA Consultancy Services (TCS) for funding support, and Dr. Rekha Singal for insightful discussion and feedback.}
}

\author{\IEEEauthorblockN{Yassine Ghannane}
\IEEEauthorblockA{\textit{Electrical and Computer Engineering} \\
\textit{Cornell University, New York, U.S.A}\\
\href{mailto:yg496@cornell.edu }{yg496@cornell.edu }}
\and
\IEEEauthorblockN{Mohamed S. Abdelfattah}
\IEEEauthorblockA{\textit{Electrical and Computer Engineering} \\
\textit{Cornell University, New York, U.S.A}\\
\href{mailto:mohamed@cornell.edu}{mohamed@cornell.edu}}
\vspace{-2cm}
}

\maketitle

\vspace{-2cm}

\begin{abstract}

Datacenters are increasingly becoming heterogeneous, and are starting to include specialized hardware for networking, video processing, and especially deep learning.
To leverage the heterogeneous compute capability of modern datacenters, we develop an approach for compiler-level partitioning of deep neural networks (DNNs) onto multiple interconnected hardware devices.
We present a general framework for heterogeneous DNN compilation, offering automatic partitioning and device mapping. 
Our scheduler integrates both an exact solver, through a mixed integer linear programming (MILP) formulation, and a modularity-based heuristic for scalability. 
Furthermore, we propose a theoretical lower bound formula for the optimal solution, which enables the assessment of the heuristic solutions' quality.
We evaluate our scheduler in optimizing both conventional DNNs and randomly-wired neural networks, subject to latency and throughput constraints, on a heterogeneous system comprised of a CPU and two distinct GPUs.
Compared to na\"ively running DNNs on the fastest GPU, he proposed framework can achieve more than 3$\times$ times lower latency and up to 2.9$\times$ higher throughput by automatically leveraging both data and model parallelism to deploy DNNs on our sample heterogeneous server node. 
Moreover, our modularity-based ``splitting" heuristic improves the solution runtime up to 395$\times$ without noticeably sacrificing solution quality compared to an exact MILP solution, and outperforms all other heuristics by 30--60\% solution quality.
Finally, our case study shows how we can extend our framework to schedule large language models across multiple heterogeneous servers by exploiting symmetry in the hardware setup.
Our code can be easily plugged in to existing frameworks, and is available at \href{https://github.com/abdelfattah-lab/diviml}{https://github.com/abdelfattah-lab/diviml}.
\end{abstract}

\section{Introduction}

Deep neural networks (DNNs)
have emerged as an important computing paradigm making significant breakthroughs in many fields.
However, DNNs are both computationally-intensive and memory-hungry, leading to a major hardware restructuring of modern datacenters to keep up with this insatiable compute demand.
GPUs are becoming commonplace, FPGAs have been included by companies like Microsoft~\cite{7783710}, and custom DNN accelerators such as Google’s TPU~\cite{10.1145/3079856.3080246} are continuously being developed. 
DNNs themselves are composed of a growing list of diverse building blocks such as convolutions, matrix-multiplications, element-wise operations, non-linear functions and shape transformations. 
Each of those primitives exhibits different vectorization patterns, sparsity and quantization tolerance and so may be suitable for implementation on different hardware accelerators \cite{dla,dynamap19}.

In addition to hardware heterogeneity, DNN topologies are becoming evermore irregular and complex thanks to their automated design through neural architecture search (NAS)~\cite{nasrl}.
NAS has demonstrated considerable success in creating DNN architectures that are highly efficient in terms of computational resource usage~\cite{mnasnet,brpnas,proxylessnas}. 
However, the irregular topologies it generates can be challenging to efficiently schedule on heterogeneous systems.
In fact, in its most simple form, with no resource usage constraints or batching, the problem of mapping and scheduling a set of tasks with dependence is a classical NP-Hard problem \cite{pinedoScheduling}.
Finding scalable and efficient methods for mapping such complex DNN computational graphs on heterogeneous systems is becoming more and more important to meet latency and throughput requirements imposed by modern DNNs and hardware platforms during inference.

Even though this scheduling problem has been previously explored in the context of traditional computing~\cite{harmonicpaper, 10.1145/2925426.2926286}, few works investigate the challenges associated with neural network models. 
In this paper, we investigate the scheduling of irregular DNN topologies onto heterogeneous hardware platforms with different latency and throughput requirements, under different batching conditions, and leveraging the \textit{module-based} nature of DNNs to significantly improve the speed and quality of our automatic scheduler.
Many have used randomly-wired neural networks (RWNNs)~\cite{RWNN} to represent NAS-designed DNNs in the context of scheduling~\cite{ordering_chaos}, and we follow suit.
Our scheduler operates on a coarse-grained computational graph of DNNs that is available through domain-specific frameworks such as PyTorch~\cite{pytorch} or TVM~\cite{tvmpaper}.
Our goal is to create a fast heterogeneous scheduling plugin that can be easily integrated into these DNN frameworks to leverage heterogeneous computing platforms.



To achieve this goal, we curate a set of DNNs from the vision domain, both manually-designed ones such as ResNet~\cite{resnet}, and NAS-found DNNs represented by an assortment of RWNNs.
We investigate the scheduling of these DNNs on a sample heterogeneous computing platform with two GPUs and a CPU, and we demonstrate a considerable improvement compared to many past heuristic baselines.
Our key algorithmic contribution is a fast DNN splitting heuristic, MILP-SPLIT, that detects and schedules each DNN module separately then combines the schedules in either an optimal or quasi-optimal fashion depending on the nature of the connection between modules.
MILP-SPLIT also comes with a theoretical lower bound for the optimal solution, which facilitates the evaluation of the scheduling quality. 
Our contributions are enumerated below: 

\begin{enumerate}
    \item We formalize the problem of partitioning and scheduling a DNN onto interconnected hardware devices in a heterogeneous computing system. We leverage both model and data parallelism to handle two core optimization objectives; latency and throughput. 
    \item We propose a novel linear mathematical programming model which is the first, up to our knowledge, scheduling problem formulation capable of handling both model and data parallelism for batched DNN execution.
    \item We introduce MILP-SPLIT: A splitting heuristic to schedule complex modular DNNs. Alongside, we perform a rigorous theoretical analysis on the implications of modularity and inter-module communication channels, on the performance of our heuristic, via the proposal of a lower bound formula.
    \item We evaluate our algorithms on computer-vision DNN benchmarks, on both mainstream DNNs and randomly wired neural networks. Compared to a single device, we achieve more than $3\times$ lower latency and $2.9\times$ higher throughput. Compared to heursitics from prior work, we achieve 30--60\% better solution quality, and up to 395$\times$ speedup compared to an exact solution.
\end{enumerate}

\section{Related Work}

On the topic of general software partitioning, there exists previous work regarding heterogeneous compilation \cite{harmonicpaper}. 
In particular, Polly-Acc offers an automatic heterogeneous compute compiler targeting CPU-GPU systems where at the compiler IR level, interesting compute kernels are detected, extracted, and modeled, and whose execution strategy is described as a schedule tree \cite{10.1145/2925426.2926286}.
AMAP is an online adaptive decision algorithm to determine if the improvement from running a function in hardware outweighs the overhead of transferring the parameters \cite{5161220}, whereas \cite{1253185}
proposes a dynamic program scheduling
approach based on the sampled energy-delay product during tentative runs.
Our approach, in contrast, is performed statically during compilation, is specifically tailored for deep learning architectures, and leverages coarse graph-level descriptions of DNNs.


Under the scope of DNN based partitioning, many existing research endeavors focus solely on training \cite{8574576,alpa}. 
Alpa automates the search for pipeline-parallel schedules for DNN training on homogeneous multi-node GPU clusters.
ParDNN introduces a graph slicing heuristic which forms primary clusters, the first iterative critical paths of the graph, and secondary clusters, the single nodes or remaining paths, and optimizes for load balancing during training~\cite{QARARYAH2021102792}. 
Chen at al.~\cite{10.1007/978-3-030-29611-7_5} propose heuristic methods to optimize latency based on Heterogeneous-Earliest-Finish-Time (HEFT) and Critical-Path for mapping and scheduling DNNs on accelerators consisting of function units such as matrix multiplication or lookup tables. 
Unlike these approaches that were specific to DNN training, our scheduling algorithm is geared towards low-latency and high-throughput inference.

Liu et al.~\cite{DenseNet} restrict their scope to the DenseNet architecture and gives an exact and efficient algorithm for its scheduling on a heterogeneous system. However, this approach is tailored for the particular topology of the DenseNet graph and is consequently difficult to generalize to broader model architectures. 
We propose a more general cut-based heuristic, which also takes advantage of the dynamic programming paradigm and can significantly speed up the mixed integer linear programming (MILP) solving.
Additionally, Mirhosein et al. \cite{RLdevice} propose a reinforcement learning approach to DNN mapping for both training and inference latency optimization. This suffers however from a lack of generalization with a need to set manually load specific parameters and with training time ranging between 12 to 27 hours. In comparison, our approach focuses on inference, handles batched inputs and strives for efficiency by leveraging modularity while maintaining some optimality guarantees.
Finally, SERENITY achieves memory-aware scheduling of irregularly wired neural networks on a single device by resorting to graph rewriting and divide-and-conquer approaches \cite{MLSYS2020_9bf31c7f}. 
We focus instead on latency and throughput optimization on multiple heterogeneous devices, taking into account each device's memory constraints. 


\begin{figure}
    \centering
    \includegraphics[scale=0.5]{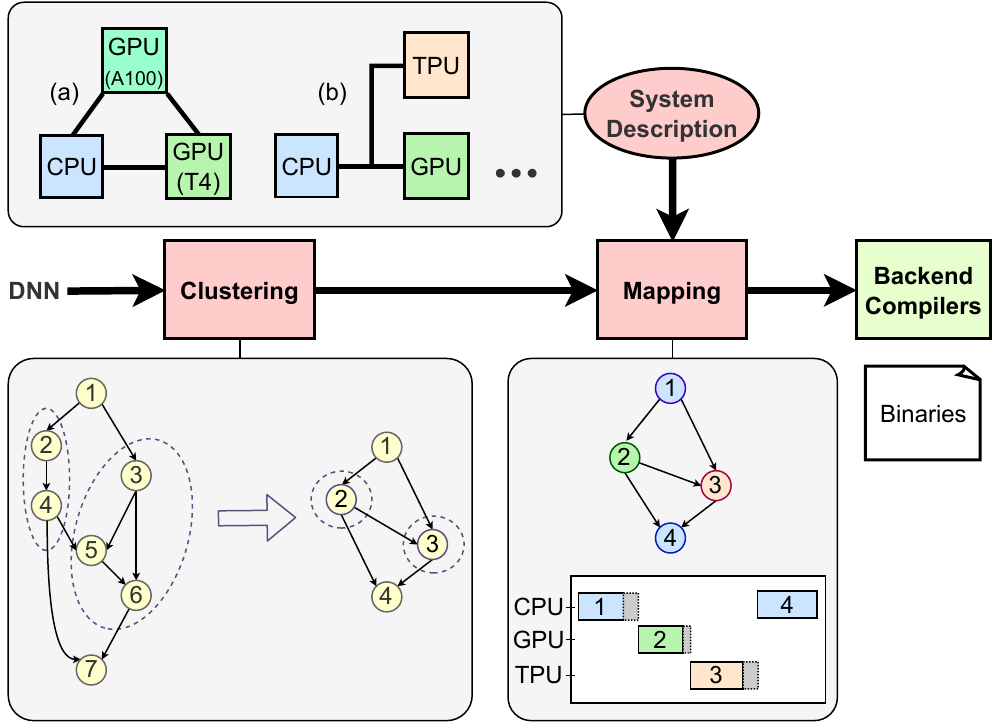}
\caption{Our heterogeneous scheduling framework.}
\label{pipeline}
\end{figure}
%

\section{Problem statement and system description}

Our approach is based on a coarse-grained representation of computational graphs that is commonly used in deep learning compilers. We present a compile-time mapping and scheduling framework for DNNs on heterogeneous hardware systems. 
The scheduler's modeling is general and agnostic to back-ends, its only limitation being what is supported by different compilers' back-ends.
Figure \ref{pipeline} illustrates how the partitioner is integrated in a DNN compilation pipeline. It is capable of reading an input consisting of a hardware system configuration and any intermediate representation (IR) of a DNN, and outputs the appropriate mapping on the system via existing compilation backends, and its corresponding schedule. 
An optional clustering step prepares the DNN graph for mapping by reducing the number of task inputs to the mapping algorithms.
A prime example is the fusion of convolution, batch normalization, and the ReLU activation function. 

   


%
%

\subsection{Problem Formulation}

We represent DNNs as a weighted directed acyclic graph (DAG), with the edges denoting data dependencies and nodes representing a DNN task (e.g. a convolutional or linear operation). 
If two tasks with data dependencies are mapped onto the same processor, the communication between them is implemented through data sharing in device memory and no communication cost is incurred. 
Each processor may execute several tasks,
but each task has to be assigned to exactly one processor, in which it is entirely executed without preemption. 
Formally, let $\mathcal{G} = (\mathcal{V}, \mathcal{E})$ be the DAG where $\mathcal{V}$ denotes the set of tasks and $\mathcal{E}$ represents the set of edges. 
Each edge $(i, j) \in \mathcal{E}$ defines a precedence relation between the tasks $i, j \in \mathcal{V}$, and is weighted by the size of the source task's output. 
A task cannot be executed unless all of its predecessors (parents) have been processed and all relevant data is available. 
Each task $i \in \mathcal{V}$ is assigned the following constants: $(wm_i)$ the data size allocated for the DNN task weights, $(im_i)$ the input tensor size and $(om_i)$ the output tensor's size.
As for our hardware system on which we map the DNN, we model it as a tuple of sets $\mathcal{H} = (\mathcal{K}, \mathcal{M}, \beta)$. $\mathcal{K}$ denotes the set of devices in our system. The two remaining sets are descriptors of the hardware system. 
$\mathcal{M} : \mathcal{K} \rightarrow \mathbb{R}^+$ is the memory capacity for each single processor and $\beta : \mathcal{K}^2 \rightarrow \mathbb{R}^+$ the communication bandwidth between linked chips---it is null if there is no link. 
If tasks $i$ and $j$ are executed on different compute nodes $h, k$ ; $h \neq k$, and $(i, j) \in \mathcal{E}$, a communication time $om_i/\beta_{h,k}$ is incurred.

The objective of this task scheduling problem is to allocate and schedule the tasks onto the compute nodes such that the overall completion time (latency) is minimized. 
We link the dataflow graph and the hardware via a map $t:(\mathcal{V},\mathcal{K}) \rightarrow \mathbb{R}^+$, which assigns to each task and device pair its corresponding latency.
We finally add to our formulation the possibility of batching and throughput optimization. 
Hence we augment our problem description with a map $\mathcal{B}:\mathcal{K} \rightarrow 2^\mathbb{N}$ that assigns to each device the subset of batch sizes supported. $t$ now describes the latency of each possible batch of similar tasks $i \in \mathcal{V}$ for each device and is redefined as $t:\mathcal{V} \times \mathcal{K} \times \mathcal{B}(\mathcal{K}) \rightarrow \mathbb{R}^+$. 
The objective is now to find for a set of $\mathcal{L}$ graph inputs the optimal mapping and scheduling of the tasks into different batches, while respecting the dependency within a single graph and the underlying resource constraints. 
Finally, we define the notion of a schedule. 
Let $\mathcal{S}:\mathcal{V}  \times [1,\dots,\mathcal{L}] \rightarrow \mathcal{K}  \times\mathbb{R}^+$ be a map which assigns each task to a device and a starting time. 
$\mathcal{S}$ is a schedule if and only if $\mathcal{S}$ respects precedence and no overlap (no two distinct batches can overlap on the same device) criteria, i.e. for every $(i, j) \in \mathcal{E}$, $l \in [1,\dots,\mathcal{L}]$: 
\begin{align*}
\mathcal{S}(i,l)_2 + 1_{ \mathcal{S}(i,l)_1 \neq \mathcal{S}(j,l)_1} \cdot m_i/\beta_{h,k} \leq \mathcal{S}(j,l)_2
\end{align*}
%
The problem statement becomes:

\begin{mdframed}
\begin{tabular}{l@{\hspace{\tabcolsep}}p{6cm}}
\multicolumn{2}{c}{\textbf{Mapping and Scheduling problem}}\\
\textbf{Input}    & Objective function (latency/throughput) $f$, $\mathcal{G}=(\mathcal{V},\mathcal{E})$, $\mathcal{S} = (\mathcal{K}, \mathcal{M}, \beta)$, $t$, $\mathcal{B}$, $\mathcal{L}$.\\
\textbf{Output}   & A schedule $\mathcal{S}:\mathcal{V}  \times [1,\dots,\mathcal{L}] \rightarrow \mathcal{K}  \times\mathbb{\mathbb{R}^+}$ which optimizes $f$
\end{tabular}
\end{mdframed}
 

\section{Algorithmic approaches}
In this section, we demonstrate our exact scheduling approach based on solving an MILP problem. 
Linear programming has been effective in solving communication constrained DAG scheduling problems for tractable instances \cite{scheduling}.
Our contributions for the exact MILP formulation are twofold: First, we incorporate memory and batching constraints into our formulation, which are commonly encountered in deep learning workloads, and we integrate our scheduler into a graph partitioning routine that we rigorously analyze to ensure the quality of its results.
However, the problem of scheduling DNNs is NP-Hard, making it intractable to find exact solutions for large graph sizes. 
Our second contribution addresses this issue. 
We take advantage of the inherent modularity in most DNNs to create fast solving schemes that are either optimal or provide strong approximation guarantees.


\subsection{MILP Problem Representation}
We introduce a novel formulation of the problem as an MILP model that explicitly considers the option of batching, where a device can process multiple inputs simultaneously. 
By incorporating batching, our formulation is better suited to capture the characteristics of modern deep learning workloads, which often involve a large numbers of inputs that can be efficiently processed in batches. 
Our approach enables us to find optimal solutions that balance the trade-offs between computation and communication costs while respecting batching and memory constraints. 
We add to the notation introduced earlier the following binary decision variables: $x_{i,j,l}$ which encodes if the DNN task $i$ corresponding to the $l$-th input is mapped to a device $j$. 
Meanwhile, $b_{i,j,l}$ describes if tasks of kind $i$ running on $j$ form a batch of size $l$, and $d_{i_1,i_2,l_1,l_2} = 1$ iff task $i_1$ from input $l_1$ is scheduled before $i_2$ from input $l_2$. 
We also consider the continuous variables: $s_{i,j}$ the starting time for processing the batch of $i$ tasks on $j$, and $C$ the total latency. 
The objective function $f$ is equal to $C$ in the latency optimization scenario or $\mathcal{L}/C$ when optimizing for throughput.
Now, we can write the mixed integer linear program, with objective \texttt{minimize C}, 
and whose constraints are as follows:

\noindent Condition \ref{eq1} asserts that each task is assigned to a single machine:
\begin{align}
\sum\limits_{u \in \mathcal{K}}x_{i,u,l} = 1; \;\;   i \in \mathcal{V}, \;\; l = 1,\dots,\mathcal{L} \label{eq1}
\end{align}
Condition \ref{eq2} ensures that each task finishes within the reported latency : 
\begin{align}
s_{i,u} + \sum\limits_{l \in \mathcal{B}_u}b_{i,u,l} \cdot t_{i,u,l} \leq C; \;\;  i \in \mathcal{V},\;\; u \in \mathcal{K} \label{eq2}
\end{align}
Condition \ref{eq3} is the condition expressing the dependency and communication constraint:
\begin{equation}
\begin{split}
 s_{i,u} + \sum\limits_{p \in \mathcal{B}_u}b_{i,u,p} \cdot t_{i,u,p} + (om_{i}/\beta_{u,v})\cdot (x_{j,v,l} + x_{i,u,l} - 1) \\ 
 \leq s_{j,v}; \;\; j \in \mathcal{V},\;\; i \in par(i),\;\; u,v \in \mathcal{K},\;\; l 
 =1,\dots, \mathcal{L} \label{eq3}
 \end{split}
\end{equation}
Condition \ref{eq4} ensures that the batch decomposition adds up correctly to the total number of items in the batch:
\begin{align}
\sum\limits_{u \in \mathcal{K}}\sum\limits_{l \in \mathcal{B}_u}l \cdot b_{i,u,l} = \mathcal{L}; \;\; i \in \mathcal{V} \label{eq4}
\end{align}
The following condition \ref{eq5} ensures that only supported batch sizes are chosen:
\begin{equation}
\begin{split}
b_{i,u,l} = 1 \; \; \; \text{iff} \; \sum\limits_{l' \in [1 \dots \mathcal{L}]}x_{i,u,l'} = l; \\ 
\;\;  i \in \mathcal{V},\;\; u \in \mathcal{K}, \;\; l = 1,\dots,\mathcal{L} \label{eq5}
\end{split}
\end{equation}
In its form above, it is not a linear equation but we can linearize it via the \textsc{big M method} \cite{Cococcioni2021}.

Condition \ref{eq6} holds the memory constraint under the supposition that all data should be preemptively moved: 
\begin{equation}
\begin{split}
\sum\limits_{i \in \mathcal{V} }((im_{i}+om_{i})\sum\limits_{l \in [1 \dots \mathcal{L}]}x_{i,u,l} + wm_{i} \sum\limits_{l \in \mathcal{B}_u}b_{i,u,l}) \\ \leq \mathcal{M}_u;  u \in \mathcal{K} \label{eq6}
\end{split}
\end{equation}

Conditions \ref{eq7} ensures no overlap of device usage between different batches. We linearize it similarly to condition \ref{eq5}:

\begin{equation}
\begin{cases}
s_{i,u} + \sum\limits_{p \in \mathcal{B}_u}b_{i,u,p} \cdot t_{i,u,p} - s_{j,u}  \leq 0 \\
\text{or} \\
s_{j,u} + \sum\limits_{p \in \mathcal{B}_u}b_{i,u,p} \cdot t_{i,u,p} - s_{i,u} \leq 0 \label{eq7}
\end{cases}
\end{equation}
\vspace{-.7cm}
\begin{align*}
\text{if} \;\; x_{i,u,l_1} = x_{j,v,l_2} = 1 ;
\\  i,\;j \in \mathcal{V},\;\; u \in \mathcal{K},\;\; i \neq j, l_1,\;l_2 = 1,\dots, \mathcal{L} 
\end{align*}


An optimization of the formulation of the MILP is to restrict constraint \ref{eq7} to pairs of tasks (i, $l_1$) and (j, $l_2$) which do not belong to the same batch graph or are not part of a path in the DAG. 
The system remains equivalent to the original as the other constraints from \ref{eq7} are enforced by the dependency constraint \ref{eq3}. 
Eliminating these redundant constraints is done by computing the transitive closure of the graph and which can be obtained efficiently with Purdom's algorithm \cite{Purdom1970ATC}.

\subsection{MILP-SPLIT: Leveraging Graph Modularity}
\subsubsection{Single-channel modularity}
\label{sec:milp_split}
\begin{figure}
  \centering

    \begin{subfigure}[b]{\linewidth}
    \centering
    \includegraphics[width=\linewidth]{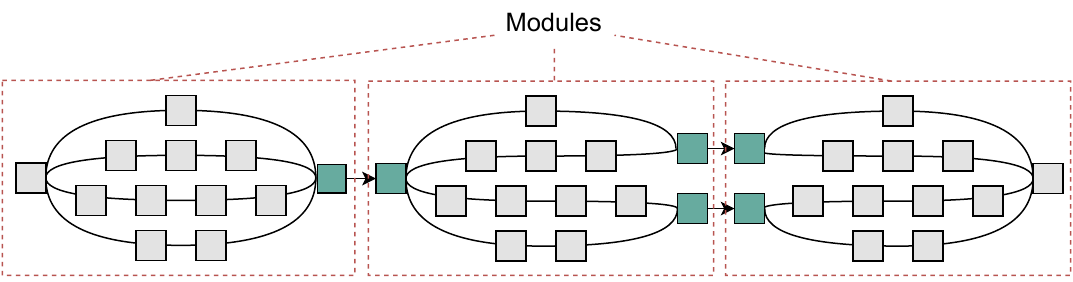}
    \caption{Partitioning scheme on modular graph}
    \label{fig:figure1}
  \end{subfigure}
  
 \vspace{0.1cm}
 
  \begin{subfigure}[b]{0.62\linewidth}
    \centering
    \includegraphics[width=\linewidth]{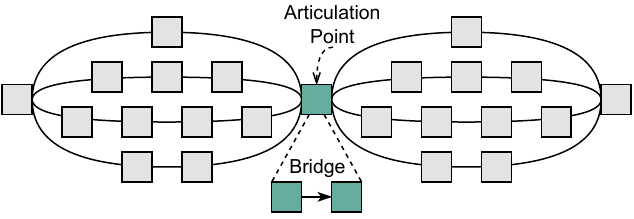}
    \caption{Articulation point and bridge between Inception modules.}
    \label{fig:articbridge}
  \end{subfigure}
  \hfill
  \begin{subfigure}[b]{0.35\linewidth}
    \centering
    \includegraphics[width=\linewidth]{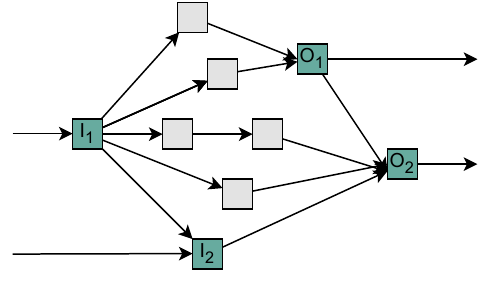}
    \caption{Example of 2-channel Erdos-Renyi sdep module}
    \label{fig:sdep}
  \end{subfigure}
  \vspace{0.5cm}
  \begin{subfigure}[b]{0.62\linewidth}
    \centering
    \includegraphics[width=\linewidth]{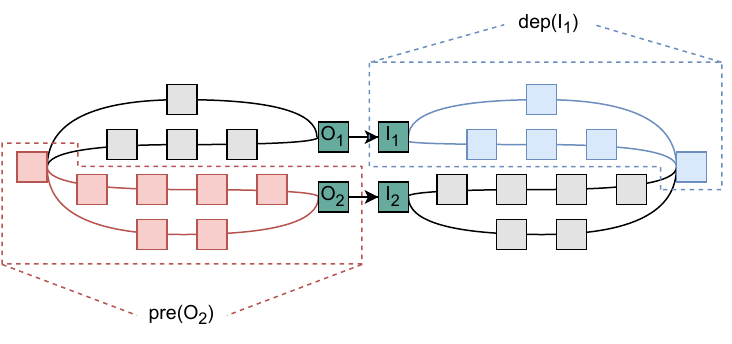}
    \caption{Dependency and predecessor subgraphs on channel endpoints}
    \label{fig:deppre}
  \end{subfigure}
  \hfill
  \begin{subfigure}[b]{0.35\linewidth}
    \centering
    \includegraphics[width=\linewidth]{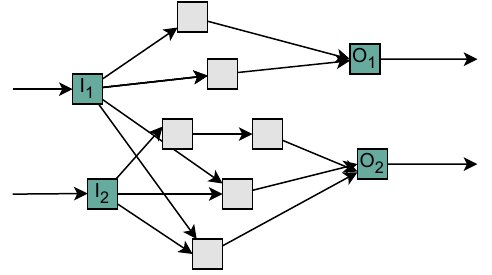}
    \caption{Example of 2-channel Erdos-Renyi wdep module}
    \label{fig:wdep}
  \end{subfigure}
  \vspace{-.3cm}
  \caption{Modularity in DNN graphs. \textbf{sdep} : all paths within a module stem from (converge toward) at least one input (output). \textbf{wdep} : module inputs and outputs are randomly sampled for their dependencies.}
  \label{fig:foursubfigures}
\end{figure}

The presence of highly connected clusters is a prevalent feature in many DNN graph structures. 
An example is shown in Figure~\ref{fig:figure1}
This characteristic can be leveraged by the scheduler to partition the global problem into independent sub-problems consisting of weakly communicating modules. This approach is particularly useful when dealing with graphs that consist of modules linked to one another, such as ResNets~\cite{resnet}, Inception~\cite{inception}, or especially RWNNs~\cite{RWNN} that are composed of several instances of sequentially linked random graph modules.

A straightforward method to identify these modules involves detecting articulation points or bridges in the graph, which correspond to vertices or edges whose removal disconnects the undirected graph, grouping tasks between them into the same module, and solving each subproblem independently.  
However, this approach can lead to suboptimal solutions as it does not account for communication costs through bridges and may result in inconsistent assignments of articulation points across modules. Fortunately, a dynamic programming solution exists to address these issues.
To obtain an optimal global solution for the whole graph, we compute the optimal schedule for each module for every possible input-device and output-device pairings, and we combine the resulting building blocks into the best configuration. 
As a preprocessing step, we transform articulation points that are not endpoints of bridges into bridge edges by introducing a dummy node and a zero-cost edge between them. 
We also add an additional constraint that mandates the mapping of these two vertices to the same device in the global solution as is illustrated in Figure \ref{fig:articbridge}. 
From now on, we refer to bridges as ``communication channels". 

Formally, Let $\mathcal{G}(\mathcal{V}, \mathcal{E})$ be a DAG with single input and output. 
We denote by $\mathcal{I}(\mathcal{Q},\mathcal{F})$ the graph obtained by reducing every module into a single vertex, where $\mathcal{Q}$ is a partition of $\mathcal{V}$ into a set of disjoint modules and $\mathcal{F} := \{(u,v) \in \mathcal{Q}^2 | \;\;\exists x \in u \; \exists y \in v \;\; (x,y)\in \mathcal{E}\}$. 
In particular, if $\mathcal{Q}$ is defined as the set of vertex modules, then $\mathcal{I}$ is a path, and we can enumerate $\mathcal{Q}$ as the set $[1, \dots, |\mathcal{Q}|]$, and through this ordering we can obtain a dynamic programming problem formulation. 
For a given module $M_t \in \mathcal{Q}$ and a pair of devices $u, v\in \mathcal{K}$ onto which the input and output of $M_t$ are mapped, and if we denote by $opt$ the solution of a module subproblem, the recursion can be written as: 
\begin{equation*}
\begin{split}
    dp(M_t,u,v) &= min_{\substack{u',v' \in \mathcal{K}}}\Big(dp(M_{t-1},u',v') \\ &+ com(t, v',u)\Big) + OPT(M_t,u,v) 
\end{split}
\end{equation*}

The effectiveness of the proposed splitting method is influenced by the number and size balance of the extracted modules. The complexity of the approach can be expressed as $O(|\mathcal{K}|^2|\mathcal{Q}|\mathbb{T})$, where $\mathbb{T}$ represents a runtime bound for each module. This complexity analysis assumes a specific cutting strategy, but can be generalized to arbitrary cuts, where $\mathcal{I}$ becomes a multigraph. 


\subsubsection{Multi-channel modularity} 

Modularity is an important property of graphs that enables exact solving for the scheduling problem on large graphs using a divide-and-conquer approach. 
However, many graphs can not be split into distinct modules of comparable size that communicate through a \textit{single} input-output channel. 
In such cases, it may still be possible to decompose the graph into balanced modules that communicate through \textit{multiple} edges, and solve for each subgraph independently. 
Figure~\ref{fig:figure1} shows an example with 1 and 2 channels. Identifying the modules boils down to computing the $k-$edge connected components \cite{10.1145/2463676.2465323} where $k-1$ is the number of channels.
Although this approach may result in a loss of optimality, it can significantly improve runtime without a significant reduction in quality.
In the case of partitioning a large graph into multichannel communicating modules, it is desirable to compute a lower bound on the optimal solution to evaluate the quality of the MILP-SPLIT (or other) heuristic, especially when solving for the entire graph is not tractable. 
In order to express the lower bound for a DAG $\mathcal{G}(\mathcal{V},\mathcal{E})$ that can be split into multichannel communicating modules, we first define for a fixed $T\subseteq\mathcal{V}$ and for every node $u$ in $\mathcal{G}$ the set of nodes $dep(u)_T = \{v \in T \;|\; \text{there is a path from $u$ to $v$}\}$, which we will refer to as the dependency set of $u$, and the set of nodes $pre(u)_T = \{v \in T \;|\; \text{there is a path from $v$ to $u$}\}$, and which we will refer to as the predecessor set of $u$ (as shown in Figure \ref{fig:deppre}). 
Let $M_1, \dots, M_{|\mathcal{Q}|}$ be a decomposition of $\mathcal{G}$ into such modules, where $\bigcup_{1 \leq t \leq {|\mathcal{Q}|}} M_t = \mathcal{V}$. 
We denote by $\mathcal{G}_s = \bigcup_{s \leq t \leq {|\mathcal{Q}|}} M_t$. 
Our approach is to proceed inductively by computing the lower bound in a recursive manner, and using the following remark:
\begin{remark}
Let $c$ denote the number of channels, and $(I_t)_{1 \leq t \leq c}$ and $(O_t)_{1 \leq t \leq c}$ denote respectively the set of vertices in the communication channels between $M_1$ and $\mathcal{G}_2$ for which the edges are in-going and out-going, i.e., the inputs of $\mathcal{G}_2$ and the outputs of $M_1$. 
For any valid scheduling of the whole graph, there exists a $t'$ such that the subgraph induced on $dep(I_{t'})_{\mathcal{G}_2}$ is completely scheduled after $M_1$, and there exists a $t"$ such that $pre(O_{t"})_{M_1}$ is completely scheduled before $\mathcal{G}_2$.
\end{remark}
Hence, if we denote by $OPT$ the function mapping subgraphs of $\mathcal{G}$ onto their optimal schedule, then we obtain the pair of inequalities: $$OPT(\mathcal{V}) \geq OPT(M_1) + min_{u \in \text{inputs}}(OPT(dep(I_u)_{\mathcal{G}_2}))$$ and $$OPT(\mathcal{V}) \geq OPT(\mathcal{G}_2) + min_{v \in \text{outputs}}(OPT(pre(O_v)_{M_1}))$$ 
The lower bound of the problem is obtained as the maximum value among the right-hand sides of the inequalities. 
This lower bound can be immediately extended to the batched throughput scenario by observing that the partial ordering defined earlier for dependency, predecessor, and module subgraphs applies to scheduling the minimal batch size that can be executed on each device. 
Specifically, it is necessary to schedule a minimum portion of the input to maintain the specified constraints via the communication channels outlined in the remark. 
However, we can do better; let $M_1$ and $dep(I_{t'})_{\mathcal{G}_2}$ be defined as in the remark; then if $\mathcal{L}$ is the total input batch to be processed and $b$ any batch size supported on every device, then there is at least a batch of $\mathcal{L}-b+1$ that needs to be processed through $dep(I_{t'})_{\mathcal{G}_2}$ after scheduling a load $b$ of $M_1$. The same reasoning holds between $OPT(pre(O_v)_{M_1})$ and $\mathcal{G}_2$, and recursively throughout the graph. 
These bound computations can be accomplished efficiently using the presented recursive formula, which lends itself well to parallelization due to the independent nature of the subproblems considered.

\section{Evaluation}\label{Evaluation}

We evaluate our mapping and scheduling framework on mainstream DNN models, a set of computer vision neural networks popular in the field of image classification, from the \textit{Torchvision} model library, and on randomly wired neural networks (RWNNs) also performing image classification tasks~\cite{RWNN}. 
We focus more on the latter because the topological irregularity of RWNNs makes it more difficult to have a good intuition on what a good mapping and scheduling should look like thus necessitating automated algorithms. 
We choose representatives from three random graph models (Erdos-Renyi, Watts-Strogatz and Barbasi-Albert), with parameters chosen corresponding to the seeds which achieved the best accuracy in prior work~\cite{RWNN}: we sample 6 models generated with parameters WS(4, 0.75), ER(0.2) and BA(5), and with module size $N\in \{10,32\}$.
We consider systems comprised of a CPU (Intel Xeon (skylake) CPU 2.00GHz) and two different GPUs (Nvidia Tesla T4 and A100 GPUs) connected by a 16 lanes PCIe 4.0 link to represent a typical heterogeneous system---relative speeds are shown in Table~\ref{tab:device_speeds}. The complete pipeline of our scheduler's evaluation setup for the aforementioned networks starts with a Pytorch model. To convert it into a coarse grain DAG, we use the torch.fx \cite{10.48550.2112.08429} symbolic tracer and in particular the Interpreter class. This class is responsible for executing an FX graph, which represents the dataflow graph of DNN inference on a node-by-node basis. By overriding the node run method, we can individually measure the performance of executing each computational node on different backends by invoking the appropriate routine on the respective node, thus creating our DAG while simultaneously benchmarking each operation on every device.

Our experiments perform a thorough comparison of our exact MILP solution, our modularity-based splitting heuristic (MILP-SPLIT), and a large number of established baselines from prior work, introduced in Section~\ref{sec:results:baselines}.
We present our findings when optimizing solely for latency (Section~\ref{sec:results:latency}) using model parallelism, and when optimizing for throughput (Section~\ref{sec:results:throughput}) using both data and model parallelism.
In both cases, we evaluate the solution quality and cost for Torchvision models, for single-module RWNNs, and for multi-module RWNNs.
Our findings demonstrate the superiority and practicality of MILP-SPLIT compared to existing baseline algorithms, and the fidelity of our estimated lower bound. 
\begin{figure*}
\fbox{
\parbox{\dimexpr\textwidth - 2\fboxsep - 2\fboxrule}{
\centering

\begin{minipage}[b]{\textwidth}
\centering
\bf Latency Optimization
\vspace{.1cm}
\end{minipage}

\begin{minipage}[b]{0.48\textwidth}
\includegraphics[width=\textwidth, trim=0 .4cm 0 0]{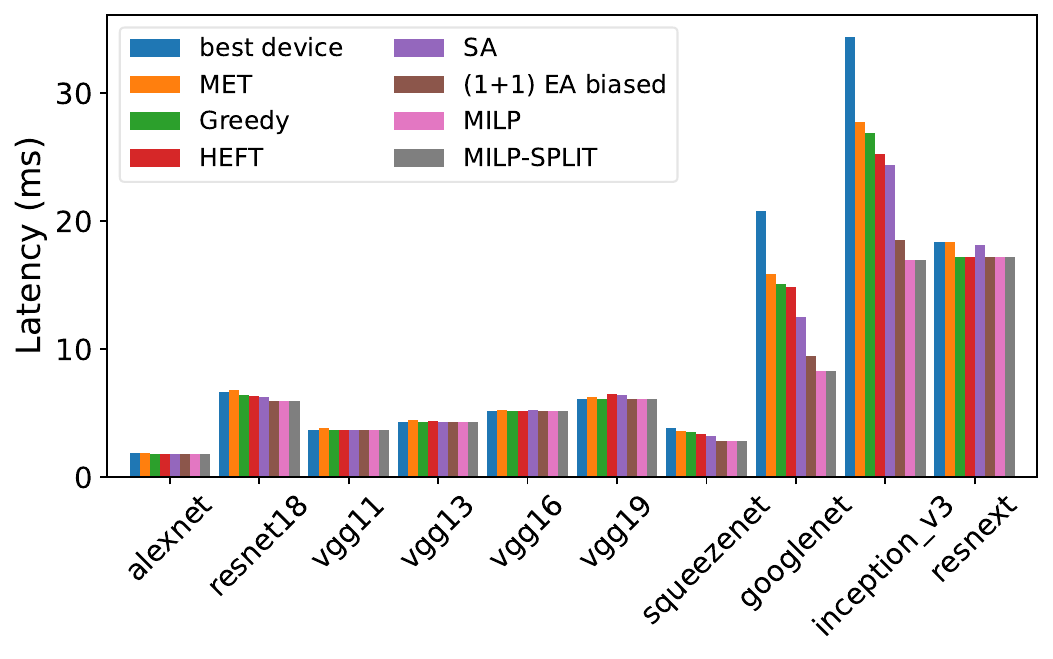}
\captionsetup{justification=centering}
\caption{Inference latency for Torchvision DNNs deployed on a heterogeneous platform with different schedulers.}
\label{fig:lattorch}
\end{minipage}
\begin{minipage}[b]{0.48\textwidth}
\includegraphics[width=\textwidth]{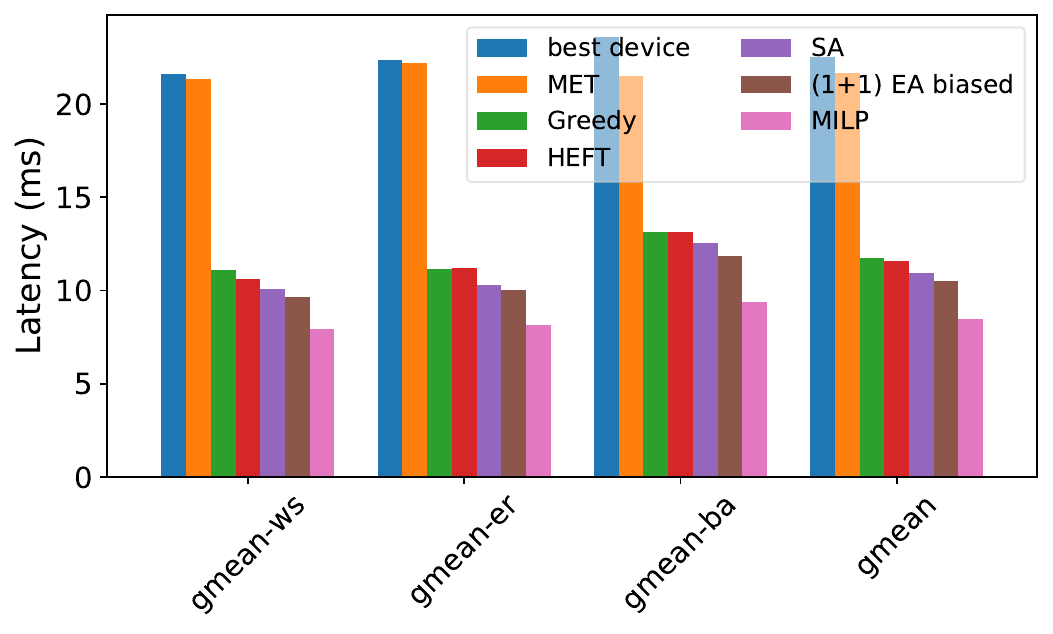}
\captionsetup{justification=centering}
\caption{Inference latency for single RWNN modules on a heterogeneous platform with different schedulers.}
\label{fig:latrwnn}
\end{minipage}

\vspace{.2cm}

\begin{minipage}[b]{\textwidth}
\centering
\small
    \captionsetup{justification=centering}
    \captionof{table}{Latency of RWNNs consisting of 10 modules. Results reported in milliseconds (ms).\\Best and second best results are highlighted in red (bold) and blue respectively.}
    \label{tab:latrwnn}
    \begin{tabular}{ccccccccc|c}
    \toprule
        \textbf{model} & \textbf{best device} & \textbf{MET} & \textbf{Greedy} & \textbf{HEFT} & \textbf{(1+1) EA biased} & \textbf{SA} & \textbf{MILP} & \textbf{MILP-SPLIT} & \textbf{LBound} \\ 
        \midrule
         1-chan & 211.7  & 209.9  & 104.4 & 99.9  & 97.5 & 98.9 & \color{red}{\textbf{80.1}} & \color{red}{\textbf{80.1}}  & 80.1 \\ 
          \midrule
        sdep, 2-chan & 234.9  & 219.3& 111.8 & 109.3  & 103.8  & 106.2  & \color{red}{\textbf{78.9}} & \color{blue}{79.1} & 73.7  \\ 
        sdep, 3-chan & 236.5  & 235.4  & 114.2 & 108.5  & 104.7  & 106.1  & \color{red}{\textbf{79.9}} & \color{blue}{80.3} & 68.1  \\ 
        sdep, 4-chan & 250.5 & 249.1  & 116.3 & 111.3 & 107.6 & 109.5 & \color{red}{\textbf{79.1}} & \color{blue}{79.4} & 61.3 \\ 
        \midrule
        wdep, 2-chan & 225.2 & 223.8 & 103.7 & 101.7 & 97.3 & 98.1 & \color{red}{\textbf{74.3}} & \color{blue}{77.6} & 73.3 \\ 
        wdep, 3-chan & 229.9 & 229.3 & 107.4 & 104.3 & 100.9 & 103.1 & \color{red}{\textbf{76.6}} & \color{blue}{78.7} & 71.8 \\ 
        wdep, 4-chan & 242.9 & 240.7 & 106.8 & 104.4 & 102.0 & 103.6 & \color{red}{\textbf{71.6}} & \color{blue}{77.0} & 62.1 \\ 
       \bottomrule
    \end{tabular}
\end{minipage}

}
}
\end{figure*}
\subsection{Baselines and Heuristics}
\label{sec:results:baselines}
We compare our MILP solver and MILP-SPLIT against popular scheduling algorithms and general purpose optimization heuristics which have shown success in DAG scheduling contexts or graph problems more generally. 

\begin{itemize}
    \item MET: the Minimum Execution Time algorithm is a list-based scheduling algorithm 
    that schedules tasks based on their minimum execution time to minimize the latency of a DAG. We extend the MET algorithm to the batched throughput optimization by selecting the best batch-device combination for each task. 

    \item Greedy: is a greedy heuristic 
    that considers the overall latency for scheduled tasks so far when scheduling the current task. 

    \item HEFT: the Heterogeneous Earliest Finish Time \cite{DUBEY2018725} algorithm is an effective approach for scheduling tasks in a heterogeneous computing environment. 
    It assigns tasks to processing nodes with different processing speeds to minimize overall execution time, using two phases to prioritize tasks based on estimated finish times.
    
    \item Simulated Annealing (SA) \cite{doi:10.1126/science.220.4598.671}: is a stochastic optimization heuristic algorithm that draws inspiration from statistical mechanics concepts 
    and has been widely used in various optimization problems, including scheduling, for example, to minimize latency \cite{SAsched, 5350229,Celik2021}. 
    
    \item Biased (1+1) EA: We implement a biased version of the (1+1) EA \cite{DROSTE200251} as an additional approximation heuristic. Also known as the random hill climbing algorithm, it is one of the most basic evolutionary algorithms but has been surprisingly efficient in practice  \cite{Borisovsky2002ASO, Carson01hill-climbingfinds}. 
    We qualify as biased the (1+1) EA when the initialisation is not randomly sampled but chosen in a greedy manner, by assigning each task to the device on which it runs fastest.
      
\end{itemize}

\textbf{Fitness function}: Here we give a succinct formulation of our problem as an objective function and an integer-string search space, which are adopted by two of our search heuristics: (1+1) EA and SA.
We encode the mapping solution as a string of integers, wherein each integer in the string signifies a distinct identifier of the device to which a node is mapped. 
The position of each integer in the string corresponds to the layers of the DNN, arranged in a breadth-first topological ordering.
Finally, the fitness function adopted for the latency (throughput) optimization problem
corresponds to the latency (throughput) of a sampled mapping with a breadth-first topological ordering.

\subsection{Latency Optimization}
\label{sec:results:latency}

Figure \ref{fig:lattorch} evaluates our scheduler to optimize latency for mainstream Torchvision models. 
There are no real improvements for DNNs with little to no parallelism, such as AlexNet or ResNet or VGG, the optimal schedule is usually the one where all tasks are mapped to the best performing device (A100 GPU). 
However, for models with higher parallelism, the improvement from MILP and MILP-SPLIT are significantly higher---more than 100\% and 150\% for Inception v3 and GoogLeNet respectively. 
Both MILP and MILP-SPLIT converge to the optimal solution for all Torchvision models without a substantial increase in runtime, thanks to the simplicity and regularity of these DNNs.

Next, we evaluate RWNNs which we expect to be a significantly more challenging workload. 
In our first experiment in Figure~\ref{fig:latrwnn}, we schedule a \textit{single} module on our heterogeneous system, optimized for latency.
Compared to simply running the RWNN module on the best device, there is a major $\sim$2$\times$ improvement in overall latency from fully-utilizing our heterogeneous system with a CPU and 2 GPUs.
When comparing across different scheduling algorithms, MILP converges to the optimal solution and is 22\%-26\% better than the best available heuristic on equivalent runtimes.
However, with RWNNs featuring multiple modules, ten in our experiment, solving MILP on the whole model is more difficult for the solver and is exponentially slower. 
This motivates the use of MILP-SPLIT for those more realistic multi-module RWNNs that are representative of DNNs created by NAS.

\begin{table}[t!]
    \centering
    \caption{Relative speed in milliseconds (ms) on experiment devices, averaged over our evaluated DNNs.}
    \begin{tabular}{cccc}
    \toprule
         & CPU & GPU (T4) & GPU (A100)\\
    \midrule
         Torchvision & 223.10 (29$\times$) & 12.16 (1.6$\times$) & 7.80 (1$\times$)\\
         RWNNs & 183.39 (7.10$\times$) & 32.58 (1.26$\times$) & 25.84 (1$\times$)	\\
    \bottomrule
    \end{tabular}
    \label{tab:device_speeds}
\end{table}

\begin{table}[t!]
\vspace{.2cm}
    \centering
    \caption{Speedup of the splitting heuristic for the latency optimization of RWNN models with [5, 10, 20] modules.}
    \begin{tabular}{c|ccc|ccc}
    \toprule
         & \multicolumn{3}{c|}{\textbf{sdep}} & \multicolumn{3}{c}{\textbf{wdep}}\\
         Modules & MILP & SPLIT& factor & MILP & SPLIT& factor\\
    \midrule
       5&82.69s&2.26s&37x&129.08s&2.45s&53x\\
       10&232.24s&4.83s&48x&271.66s&5.00s&54x\\
       20&1907.12s&13.49s&141x&5850.37s&14.81s&395x\\
    \bottomrule
    \end{tabular}
    \label{tab:runspeed}
\end{table}

To evaluate MILP-SPLIT, we stack multiple RWNN modules to represent realistic NAS-discovered models.
In this case, each module is generated using the ER(0.2) model and may include multiple communication channels to connect to the next module. 
As indicated by our lower bounds formulation (Section~\ref{sec:milp_split}), the density of nodes and edges that are accessible from the endpoints of communication edges can significantly impact the quality of the splitting heuristic and the accuracy of the corresponding lower bound. 
Therefore, we evaluate our splitting heuristic using two different scenarios for the topology of communication edges. 
In the first scenario, module inputs and outputs are randomly sampled for their dependencies, while in the second scenario, all paths within a module stem from (converge toward) at least one input (output). 
We refer to these scenarios as the ``weakly dependent" scenario (\textbf{wdep}) and the ``strongly dependent" scenario (\textbf{sdep}), respectively, and examples are shown in Figures~\ref{fig:wdep}~and~\ref{fig:sdep}.

Based on the results presented in Table ~\ref{tab:latrwnn}, it can be observed that our splitting heuristic (MILP-SPLIT) exhibits a solution that is in close proximity to the optimal solution. 
Additionally, this heuristic outperforms all other scheduling methods considered in this study by a significant margin, as it is $\sim$30\% better compared to the best heuristic baseline. 
Table \ref{tab:runspeed} highlights that the MILP-SPLIT heuristic provides a substantial improvement (37$\times$--395$\times$) in runtime compared to MILP when both scheduling algorithms reach their best solution. 
Also shown in Table~\ref{tab:latrwnn} is our lower bound (LBound), which offers a convenient means of obtaining a quick performance guarantee for the splitting heuristic. 
Our observations indicate that for the \textbf{wdep} models, the LBound is closer to the true optimum than for the \textbf{sdep} models, where it tends to be more pessimistic. 
This difference is attributed to the lower bound computation which considers complete overlap in scheduling separate paths originating from each module output. 
This is more likely to hold in an optimal schedule for the \textbf{wdep} scenario,  where the distribution of these paths is more evenly spread compared to the \textbf{sdep} scenario, where a specific endpoint's emanating paths cover all the predecessor or dependency subgraphs---this phenomenon is also the reason why MILP-SPLIT is closer to the optimum on \textbf{sdep} graphs.
Our results show that MILP-SPLIT is a viable and high-quality heuristic that offers lower-bound guarantees on quality.

\subsection{Throughput Optimization}
\label{sec:results:throughput}
\begin{figure*}
\fbox{
\parbox{\dimexpr\textwidth - 2\fboxsep - 2\fboxrule}{
\centering

\begin{minipage}[b]{\textwidth}
\centering
\textbf{Throughput Optimization} 

{\footnotesize(2 hours timeout for MILP and 600 seconds timeout for MILP-SPLIT.)}
\vspace{-.2cm}
\end{minipage}

\begin{minipage}[b]{0.48\textwidth}
\vspace{.5cm}
\includegraphics[width=\textwidth, trim=0 .4cm 0 0]{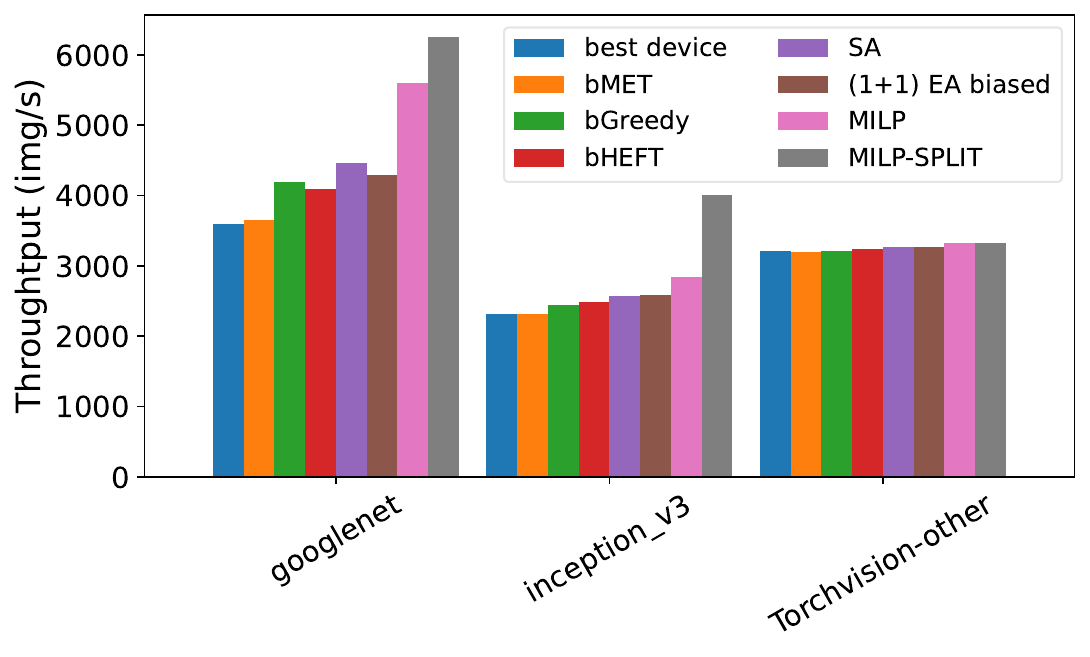}
\captionsetup{justification=centering}
\caption{Inference throughput for a batch (B=128) inputs on Torchvision models on a heterogeneous platform.}
\label{fig:tputtorch}
\end{minipage}
\begin{minipage}[b]{0.48\textwidth}
\includegraphics[width=\textwidth]{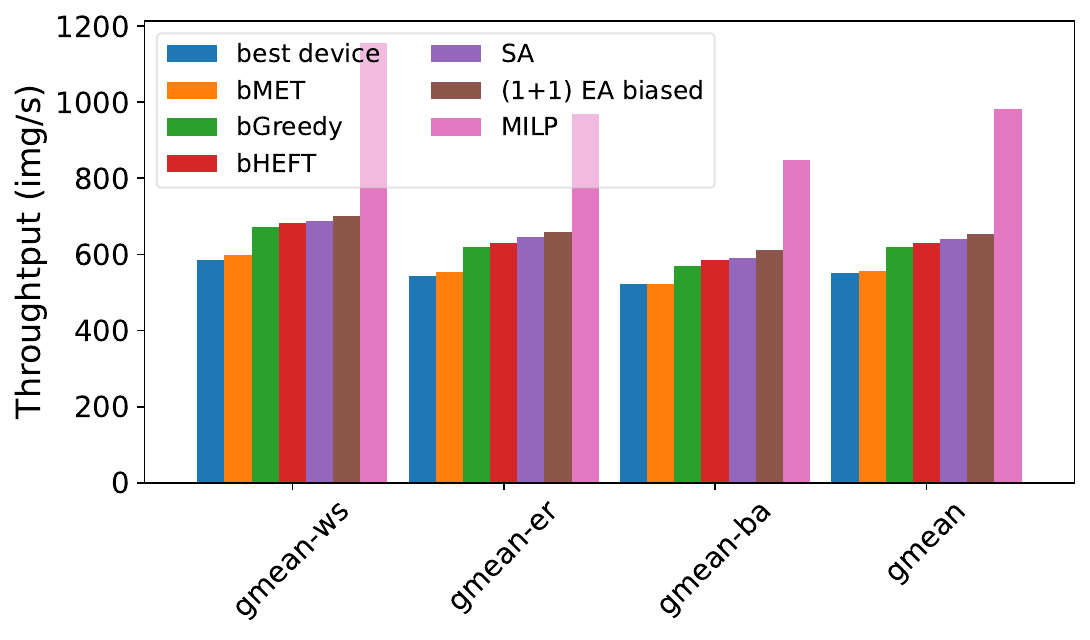}
\captionsetup{justification=centering}
\caption{Inference throughput for a batch (B=16) inputs on single RWNN modules on a heterogeneous platform.}
\label{fig:tputrwnn}
\end{minipage}

\vspace{.2cm}

\begin{minipage}[b]{\textwidth}
\centering
\small
\captionsetup{justification=centering}
\captionof{table}{Throughput for RWNNs consisting of 10 modules. Results reported in images-per-second (imgs/s).\\Best and second best results are highlighted in red (bold) and blue respectively.}
\label{tab:tputrwnn}
\begin{tabular}{ccccccccc|c}
\toprule
\textbf{Model}&\textbf{BD}&\textbf{bMET}&\textbf{bGreedy}&\textbf{bHEFT}&\textbf{(1+1) EA biased}&\textbf{SA}&\textbf{MILP} & \textbf{MILP-SPLIT} & \textbf{UBound} \\ \midrule
1-chan & 54 & 56 & 74 & 75 & 84 & 87 &  \color{blue}{114} &  \color{red}{\textbf{135}} &  164 \\
\midrule
sdep, 2-chan & 48 & 50 & 67 & 66 & 75 & 78 & \color{blue}{95} &   \color{red}{\textbf{119}} & 180 \\
sdep, 3-chan & 49 & 51 & 68 & 70 & 78 & 81 & \color{blue}{116} &   \color{red}{\textbf{129}} & 196 \\
sdep, 4-chan & 47 & 48 & 65 & 67 & 76 & \color{blue}{79} & 73 &   \color{red}{\textbf{126}} & 209 \\
\midrule
wdep, 2-chan & 51 & 53 & 76 & 75 & 86 & 87 & \color{blue}{89} &   \color{red}{\textbf{137}} & 182 \\
wdep, 3-chan & 49 & 52 & 73 & 73 & 82 & \color{blue}{85} & 72 &   \color{red}{\textbf{137}} & 181\\
wdep, 4-chan & 47 & 50 & 72 & 74 & 82 & \color{blue}{84} & 65 &   \color{red}{\textbf{138}} & 207 \\
\bottomrule
\end{tabular}
\end{minipage}

}}
\end{figure*}

\begin{figure}[h]
\centering
\includegraphics[width=\columnwidth]{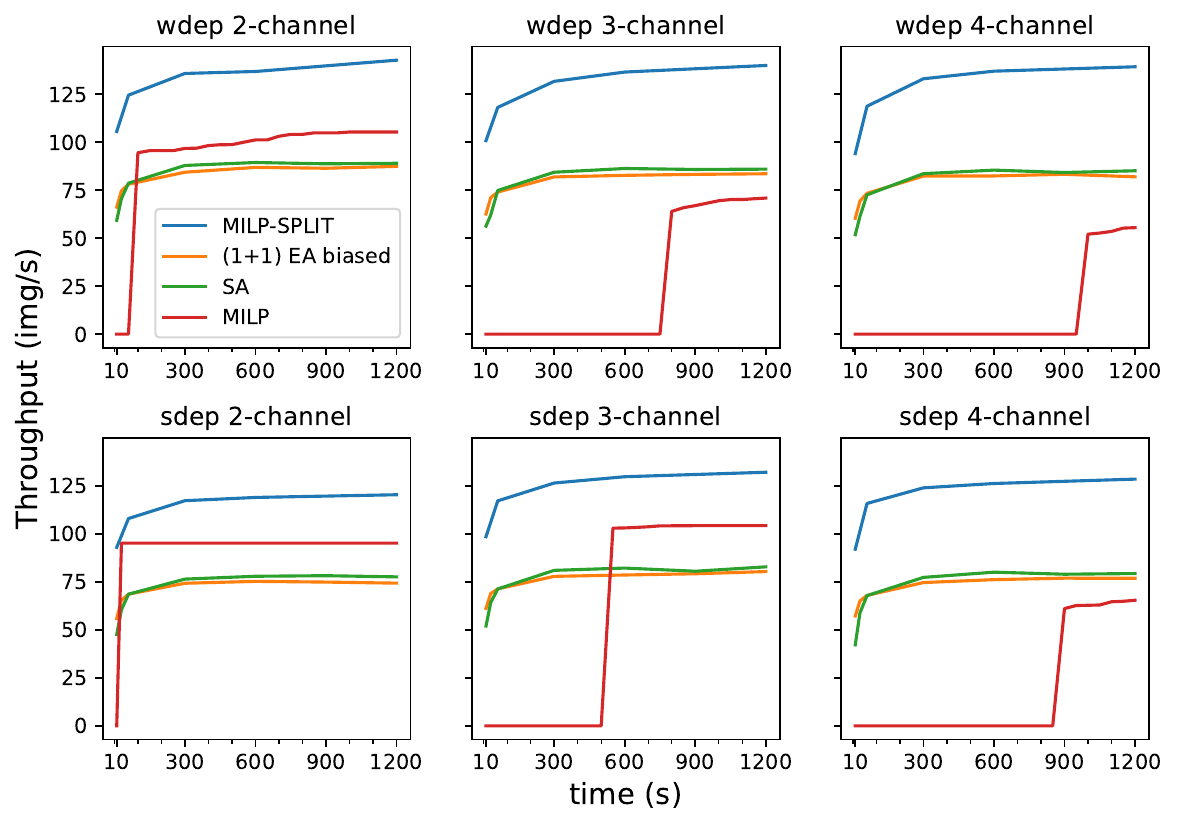}    
\caption{Solution quality (throughput) over time for MILP, MILP-SPLIT and heuristics on 10 modules RWNNs.}
\label{fig:runtimethroughput}
\end{figure}
We consider throughput optimization in the low-latency inference regime, where we batch B inputs (e.g. 128) and we find the fastest way to run that batch using the available devices. 
Successive inputs are queued together in groups of B before going to the hardware system for execution. 
This realistically mimics how inference is done in datacenters where low latency is critical to respond to user requests promptly.

Figures~\ref{fig:tputtorch}, \ref{fig:tputrwnn}, and Table \ref{tab:tputrwnn} show our throughput optimization results attained with our framework via batching. bMET, bGreedy and bHEFT are the batched equivalent of the corresponding heuristics. In this case, we have a batch of inputs B queued for processing, and our scheduler can further decompose this batch into $\sfrac{B}{4}$, $\sfrac{B}{2}$, and $\sfrac{3B}{4}$ when allocating inputs to different devices.
This enables our scheduler to leverage both model and data parallelism when mapping the DNN workload onto the hardware system. 
Unlike the latency objective, the MILP solving on the whole graph does not terminate within a 2 hours deadline, even for single RWNN modules or for regular networks with high model parallelism such as inception-based DNNs. 
Consequently, MILP-SPLIT outperforms na\"ive MILP solving both in terms of scheduling quality and runtime. 
It is worth noting that since MILP cannot reach the optimum solution for a single RWNN module, MILP-SPLIT provides only an approximate solution for each of its module schedules. 
However, our splitting heuristic achieves up to $\sim$60\% better performance than the best-performing heuristic baseline with equivalent running times. 
Results reported in Table \ref{tab:tputrwnn} are based on 600s deadlines for MILP-SPLIT and for other search heuristics, EA and SA. 
Moreover, Figure \ref{fig:runtimethroughput} provides a more detailed view of the solution quality over time, illustrating the challenge of solving the scheduling problem on the entire graph using MILP with numerous communication channels.

\comment{
Note that, in these experiments, the ``best device'' always had enough memory to run the entire batch of B inputs, and was therefore the fastest option---there is no obvious manual workload partitioning that can outperform this baseline for our latency-constrained throughput optimization.
In the following subsection, we analyze in more depth a more complex case study that further showcases the advantage of our scheduler when the fastest device can no longer run the entire batch size.
}
\begin{figure}[t]
  \centering
  \includegraphics[width=\columnwidth, trim=0 1.8cm 0 0]{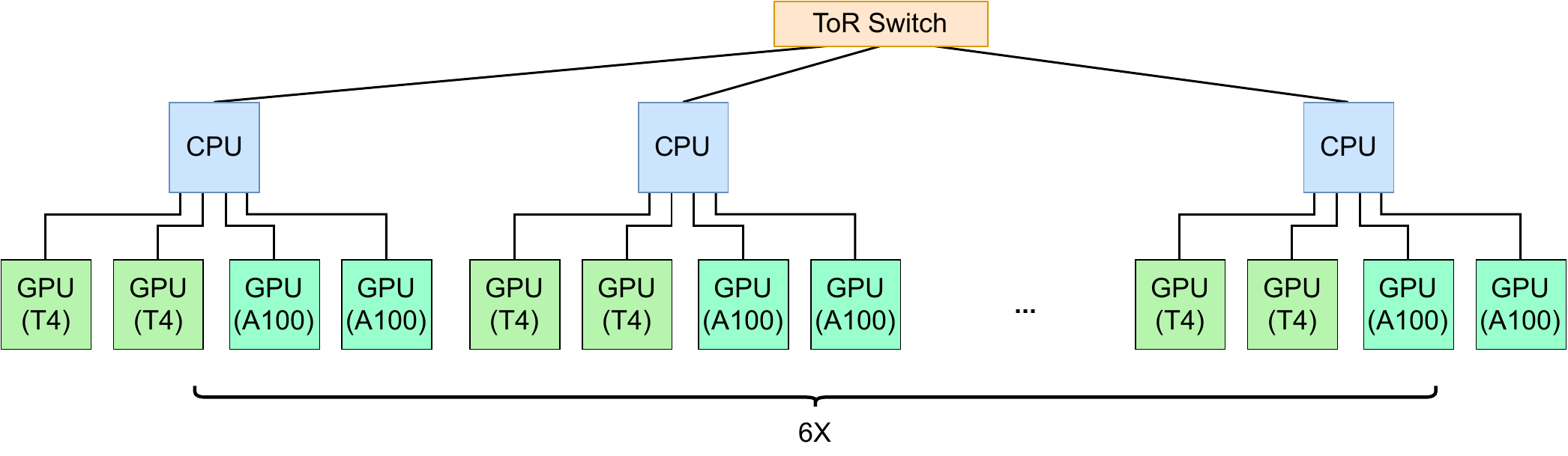}
  \caption{Multi-node heterogeneous system for GPT-3 inference.}
  \label{fig:server}
\end{figure}

\section{Case Study: GPT-3 Inference on Distributed Heterogeneous Compute Nodes}

As DNN models continue to grow in size, it has become necessary to extend our focus beyond single-node servers to extend our formulation to more complex setups.
In this case study, we investigate the use of our scheduler for a large language model (LLM), GPT-3~\cite{brown2020language}, on a distributed heterogeneous platform as shown in Figure~\ref{fig:server}. 
This model belongs to a category of deep neural networks that exhibits notable modularity, as it is predominantly constructed by stacking transformer modules \cite{vaswani2017attention}. 
In contrast to our earlier analysis of RWNNs, GPT-3 modules exhibit high regularity but the complexity of the problem stems from the larger search space of hardware device options. 
To counterbalance that, a key aspect of our analysis revolves around the exploitation of \textbf{symmetry} in the hardware platform to restrict the search space size without sacrificing solution quality.
Our preliminary results LLM scheduling only consider a single decoding step as a test workload.

As reported in Table~\ref{tab:distrib} we consider two ways to schedule our GPT-3 graph.
``Single node" utilizes our MILP solvers to schedule one-sixth of the GPT-3 graph on a single node, then replicates that schedule for the rest of GPT-3 on the remaining 5 nodes.
We consider this a competitive baseline because it automates the common practice of manually partitioning LLM inference to fit a single compute node then scaling up the number nodes.
``Multi node" exposes the entire compute system (all 6 nodes) to our MILP-SPLIT solver, but we employ \textit{symmetry-breaking} techniques to compress the solution space of the explored schedules, allowing us to find a high-quality schedule in reasonable time. 
Symmetries arise from the fact that each schedule $S$ represents a set of equivalent solutions $E_S$, where any element within this set can be derived from $S$ by permuting device mappings while maintaining the same overall latency. 
In our approach, we introduce additional constraints to our MILP formulation, enforcing a partial ordering of certain variables (e.g. \#batches, \#tasks, time utilization) between identical devices within a node or between nodes.
For example, we can ensure that the number of tasks assigned to node $i$ is always less than or equal node $j$ for $0 \leq i < j < 6$ in our example system (Fig.~\ref{fig:server}).
This retains all non-isomorphic solutions in our search space whilst compressing it by $\sim 4^66!=2.9\times10^6$, where the $6!$ and $4^6$ factors represent inter- and intra-node symmetry respectively.

Furthermore, our experimental results demonstrate that the choice of symmetry-breaking criterion can significantly impact the quality of the solution. 
This can be attributed to the phenomenon of premature convergence. 
If the symmetry-breaking constraints overly restrict the problem or generates a compressed space whose topology is not regular enough, the solver may settle for a locally optimal solution instead of exploring other potentially superior regions of the solution space either located outside of the compressed space or harder to access with the solver's intrinsic optimization heuristics due to the irregularity of the new space. 
We hypothesize that utilizing \#batches as the symmetry-breaking criterion tends to be overly restrictive, discouraging the solver from performing batch rearrangements that would contradict the ordering constraints, thus resulting in relatively smaller improvements over MILP-SPLIT without symmetries. 
On the other hand, despite the discrete nature of task variables and the continuous nature of utilization time variables, both variables are coarser grain than \#batches thus yielding comparable performance and surpassing the baseline schedule by $\sim$31\% and the single node MILP-SPLIT by $\sim$10\%. 
Our results lay the foundations towards multi-node heterogeneous scheduling leveraging MILP-SPLIT, and we aim to further explore this topic in future work.
\begin{table}
    \centering
    \caption{GPT-3 throughput (inputs/s) on our distributed system. We use a 600s timeout and each input is 20 tokens.}
    \begin{tabular}{lcc}
    \toprule
         \textbf{Heuristic} & \textbf{Throughput} \\\midrule
        Single node - MILP (baseline) & 4.8\\
        Single node - MILP-SPLIT & \color{blue}{5.7}\\
        \midrule
        Multi node - MILP-SPLIT no symmetries & 5.0\\
        Multi node -  MILP-SPLIT all symmetries – batch  &5.6\\
        Multi node -  MILP-SPLIT all symmetries – task   & \bf \textcolor{red}{6.3}\\
        Multi node -  MILP-SPLIT all symmetries – time &  \bf \textcolor{red}{6.3}\\
         \midrule
         UBound & 9.9\\
\bottomrule
    \end{tabular}
    \label{tab:distrib}
\end{table}
\vspace{-.3em}
\section{Conclusion}
We presented a general framework that leverages both data and model parallelism to schedule DNNs on heterogeneous hardware systems.
Our algorithmic approaches focused on an exact MILP solution, and a splitting heuristic, MILP-SPLIT, to utilize modularity within both conventional and randomly-wired DNNs.
Our results on both throughput and latency optimization demonstrated more than 30--60\% improvement compared to the best, and most widely-used heursitics, and MILP-SPLIT was up to $\sim$395$\times$ faster than a full MILP solution.
Finally, we extended our scheduler to larger multi-node heterogeneous server deployments by showcasing improved scheduling of GPT-3 by exploiting symmetries in the hardware system. 
In the future, we aim to expand our framework to explore more efficient methods for scheduling large DNNs on distributed systems, to handle DNN training, and to include pre- and post-processing portions of a deep learning workload.

\small
\bibliographystyle{ieeetran}
\bibliography{IEEEabrv,main}
\end{document}